\documentclass[letterpaper]{article} 
\usepackage{aaai25}  
\usepackage{times}  
\usepackage{helvet}  
\usepackage{courier}  
\usepackage[hyphens]{url}  
\usepackage{graphicx} 
\usepackage{natbib}  
\usepackage{caption} 
\usepackage{algorithm}
\usepackage{algorithmic}

\usepackage{newfloat}
\usepackage{listings}
\DeclareCaptionStyle{ruled}{labelfont=normalfont,labelsep=colon,strut=off} 
\floatstyle{ruled}
\newfloat{listing}{tb}{lst}{}
\floatname{listing}{Listing}
\usepackage{subfloat}
\usepackage{amsmath}
\usepackage{amssymb}
\usepackage{booktabs}

\usepackage{multirow}
\usepackage{bm}
\usepackage{bbold}
\usepackage{makecell}
\usepackage{xcolor}

\title{Improving Transformer Based Line Segment Detection with Matched \\ Predicting and Re-ranking}
\author{
    Xin Tong,
    Shi Peng,
    Baojie Tian,
    Yufei Guo,
    Xuhui Huang,
    Zhe Ma\textsuperscript{\rm}\thanks{Corresponding author.}
}
\affiliations{
    Intelligent Science \& Technology Academy of CASIC\\

    xin\_tong@pku.edu.cn, mazhe\_thu@163.com
}

\usepackage{bibentry}

\begin{document}

\maketitle

\begin{abstract}
Classical Transformer-based line segment detection methods have delivered impressive results. However, we observe that some accurately detected line segments are assigned low confidence scores during prediction, causing them to be ranked lower and potentially suppressed. Additionally, these models often require prolonged training periods to achieve strong performance, largely due to the necessity of bipartite matching. In this paper, we introduce RANK-LETR, a novel Transformer-based line segment detection method. Our approach leverages learnable geometric information to refine the ranking of predicted line segments by enhancing the confidence scores of high-quality predictions in a posterior verification step. We also propose a new line segment proposal method, wherein the feature point nearest to the centroid of the line segment directly predicts the location, significantly improving training efficiency and stability. Moreover, we introduce a line segment ranking loss to stabilize rankings during training, thereby enhancing the generalization capability of the model. Experimental results demonstrate that our method outperforms other Transformer-based and CNN-based approaches in prediction accuracy while requiring fewer training epochs than previous Transformer-based models.
\end{abstract}

\section{Introduction}
Line segment detection is a fundamental and critical problem in computer vision. An accurate line segment detection algorithm can significantly enable and enhance various computer vision applications, such as 3D reconstruction \cite{li2012strategy,langlois2019surface}, camera calibration \cite{zhang2016flexible,nakano2021camera}, depth estimation \cite{zavala2022depth}, scene understanding \cite{hofer2017efficient}, object detection \cite{tang2022line}, and SLAM \cite{vakhitov2019learnable,gomez2019pl}.
Traditional line segment detection algorithms directly utilize low-level information, such as image gradients, for line segment detection. While these algorithms offer fast detection speeds, they often result in fragmented line segments. In contrast, learning-based methods can detect longer and more meaningful line segments.

\begin{figure}[tb]
    \centering
    \includegraphics[width=8.3cm]{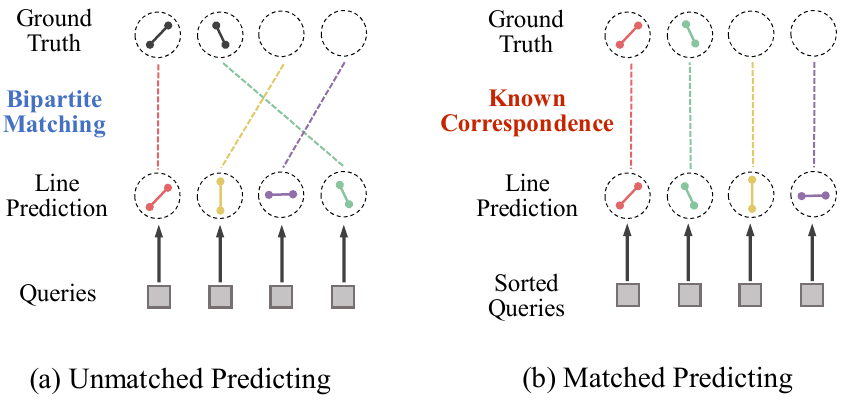}
    \caption{Comparison between unmatched and matched predicting in Transformer-based line segment detection. (a) In unmatched predicting, predictions need to be matched with ground truth online using bipartite matching. (b) In matched predicting, each prediction is directly associated with a specific ground truth value, resulting in higher training efficiency and stability.}
    \label{fig:matched_predicting}
\end{figure}

Convolutional Neural Network (CNN)-based methods typically predict the centroid of a line segment along with the offsets of its two endpoints relative to the centroid, taking advantage of the translation invariance of convolution operations. In contrast, Transformer-based methods can directly predict the two endpoints of a line segment. However, existing Transformer-based line segment detection methods employ Transformer encoder-decoder architecture, which requires online bipartite matching to define the loss function during training. As a result, these methods often demand extensive training time to converge and depend on heavy pretraining or multi-stage fine-tuning to achieve optimal results. To address this issue, we propose a novel matched prediction strategy, where the matching correspondence between predictions and ground truths is predefined, as shown in Fig.\ref{fig:matched_predicting}. Specifically, the feature point closest to the centroid of a line segment is responsible for predicting that line. The predictions are then ordered and can directly correspond to the ground truth supervision signal.

\par
Proposal-based methods typically assign a score to each predicted line segment and rank them from highest to lowest, with the top-ranked segments chosen as the final predictions. In these methods, scores and locations are predicted simultaneously, making score prediction and location regression independent processes.
However, we observe that some accurately detected line segments receive low confidence scores during prediction, leading to lower rankings and potential suppression. This observation suggests the possibility of reevaluating and optimizing the scores of predicted line segments to improve the final selection. As shown in Table \ref{tab:w/o ranking}, our experiments indicate that selecting the appropriate line segments from the candidates can significantly enhance detection performance.
To address this, we present a simple and efficient re-ranking module for line segment detection based on learnable geometric information, since low-level geometric info such as edge, endpoint, and length of line segment always provides essential insights in finding better lines. Moreover, the re-ranking module is highly interpretable and brings minimal computational overhead.

\par
In this paper, we propose a novel Transformer-based method for line segment detection. The multi-scale and multi-level image features from different layers of the CNN backbone with feature pyramid are extracted, and then processed by deformable Transformer Encoder. The scores and locations of line segments are predicted on higher-resolution feature maps. Specifically, each feature point predicts a confidence score indicating the likelihood that the centroid of a given line segment is nearest to that feature point, as well as the location of the potential line segment. To supervise the training, we directly apply confidence and position losses without the need for bipartite matching. Additionally, we introduce a novel ranking loss to ensure that feature points predicting higher-quality line segments receive higher confidence scores. Finally, the line segments are re-ranked by optimizing the scores based on learnable geometric information, with the top-ranked segments selected as the final predictions.
\par
Our contributions can be summarized as follows:

(1) We propose a novel Transformer-based line segment detection architecture where the feature point nearest to the centroid of a line segment is responsible for predicting its location. This architecture eliminates the need for bipartite matching and enables line segment detection on high-resolution feature maps.

(2) We observe the confidence scores may fail to accurately reflect the quality of the predicted line segments. To tackle the problem, we introduce a learnable geometric information based re-ranking module and a line segment ranking loss, which can significantly improve the detection performance.

(3) Experimental results demonstrate that our method outperforms other Transformer-based and CNN-based approaches in prediction accuracy while requiring fewer training epochs than previous Transformer-based models.

\section{Related Works}

\begin{table}[tb]
    \centering
    \begin{tabular}{c c c}
        \toprule
         Predictions & sAP$^{5}$(\%) & sAP$^{10}$(\%)\\
        \midrule
         w/o Re-ranking & 62.0 & 67.2 \\
         Re-ranking with GT & 91.3 (\textcolor{red}{$+$ 29.3}) & 95.5 (\textcolor{red}{$+$ 28.3})\\
        \bottomrule
    \end{tabular}
    \caption{Line detection performance before and after re-ranking using ground truth. Predicted line segments that are closer to the ground truth are assigned higher scores through re-ranking. With proper ranking, the overall performance of the predicted candidate line segments can be significantly improved. Motivated by this observation, we propose to use a re-ranking module to optimize the predicted confidence scores during inferring and use a ranking loss to supervise confidence score predicting additionally during training.}
    \label{tab:w/o ranking}
\end{table}

\textbf{Line Segment Detection.} Traditional line detection methods such as \cite{von2008lsd,akinlar2011edlines,lu2015cannylines} rely on low-level image features, e.g., image gradients. Based on local edge features, Hough transform is used for line segment detection in \cite{guil1995fast,furukawa2003accurate,xu2015closed}.
Recently, learning-based methods have achieved promising results and can be roughly divided into two categories. In junction-based methods, DWP \cite{huang2018learning} predicts junction map and edge map in two branches before merging them. PPGNet \cite{zhang2019ppgnet} uses a point-pair graph to describe junctions and line segments. L-CNN \cite{zhou2019end} applies line proposal and LoI pooling to propose candidate lines and verify them. LETR \cite{xu2021line} models it as object detection and predicts line segments with DETR architecture. 
Methods with dense prediction first predict representation map and extract line segments with post-processing. AFM \cite{xue2019learning} proposes attraction field maps to represent the image space and uses a squeeze module to generate line segment maps. HAWP \cite{xue2020holistically} further builds a hybrid model considering AFM and L-CNN. Lin \textit{et al.} \cite{lin2020deep} apply deep Hough transform to the previous detection architectures. TP-LSD \cite{huang2020tp} introduces tri-points line segment representation for end-to-end detection. M-LSD \cite{gu2022towards} presents SoL augmentation and designs an extremely efficient architecture for fast detection.
In this work, we propose a novel Transformer-based line detection method which can get proposals from dense feature maps without bipartite matching.

\begin{figure*}[t]
    \centering
    \includegraphics[width=17.7cm]{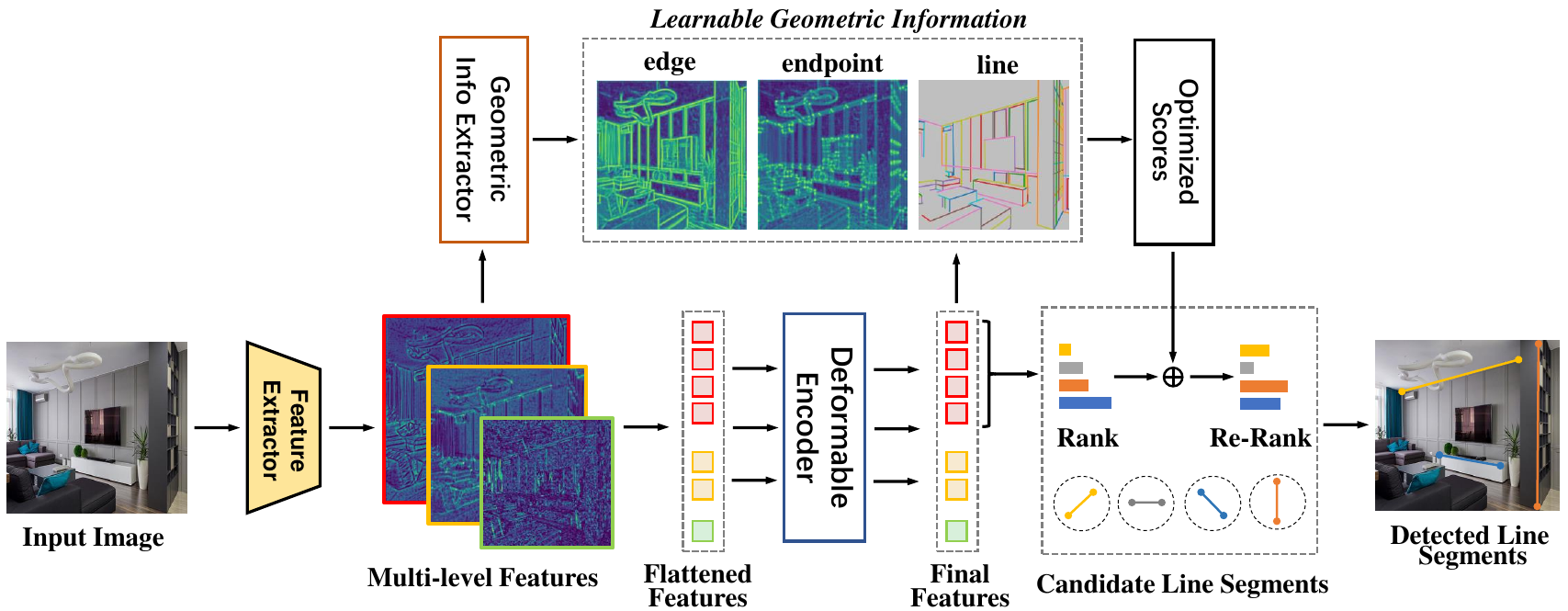}
    \caption{Overview of the proposed RANK-LETR. The process begins by feeding an image into a CNN backbone to extract multi-level feature maps from different layers. These features are then processed by a deformable Transformer encoder to generate candidate line segments. The candidate segments are predicted using high-resolution feature maps for higher prediction accuracy and less ambiguity, with each segment represented by confidence scores and positions. Each feature point is responsible for detecting the line segment whose centroid is nearest to it. Additionally, learnable geometric information is extracted from the multi-level features using a CNN-based geometric information extractor. Finally, the line segments are re-ranked by optimizing their confidence scores with the learnable geometric information.}
    \label{fig:pipeline}
\end{figure*}

\noindent\textbf{Visual Transformer.} 
Transformer-based models have gained significant success in computer vision tasks. Dosovitskiy \textit{et al.} \cite{dosovitskiy2020image} propose the Vision Transformer (ViT) which applies transformers to image classification by dividing the image into fixed-size patches. Carion \textit{et al.} \cite{carion2020end} introduce transformers into object detection pipeline through bipartite matching, named DETR. Zhu \textit{et al.} \cite{zhu2020deformable} further propose deformable DETR whose attention modules only attend to a small set of keys. Xu \textit{et al.} \cite{xu2021line} apply DETR architecture in line segment detection with a multi-scale encoder-decoder strategy. 
Tong \textit{et al.} \cite{tong2022transformer} use Transformer to cluster line segments corresponding with the same vanishing points and further apply Transformer in end-to-end vanishing point detection \cite{tong2024end}. Transformers are also used in semantic segmentation \cite{xie2021segformer}, pose estimation \cite{huang2020hand}, tracking \cite{chen2021transformer}, etc. Our work aims at improving transformer-based line segment detection with matched predicting strategy and presented re-ranking module.

\noindent\textbf{Ranking-based Losses for Object Detectors.} Ranking-based losses have been widely used recently. Average Precision Loss is first proposed \cite{chen2019towards} to address the imbalance of foreground-background classification problem by framing object detection as a ranking task. Oksuz \textit{et al.} \cite{oksuz2021rank} propose Rank \& Sort (RS) Loss that defines a ranking objective between positives and negatives as well as a sorting objective to prioritize positives with respect to their continuous IoUs.
Yavuz \textit{et al.} \cite{yavuz2024bucketed} apply Bucketed Ranking-based(BR) Losses which group negative predictions into several buckets.
Cetinkaya \textit{et al.} \cite{Cetinkaya_2024_CVPR} extend Ranking-based Loss to edge detection. In this work, we propose a line segment ranking loss for ranking the feature point with a higher line segment detection quality a higher confidence score.

\section{Method}
The overview of our algorithm is depicted in Fig. \ref{fig:pipeline}. We introduce our method from three aspects including problem modeling, network architecture and training supervision.

\subsection{Line Segment Proposal for Matched Predicting}
In traditional DETR frameworks, bipartite matching is used to assign predictions to ground truth sequences. However, these approaches can be unstable in convergence, particularly during the early stages of training \cite{li2022dn}. For the task of line segment detection, we observe that the centroids of individual line segments often do not coincide. Based on this observation, we introduce a novel centroid-based representation for line segments in Transformer-based predictions. Specifically, in proposal-based line segment detection, each feature point in the feature map has a unique 2D coordinate, and the feature point closest to the centroid of a line segment is responsible for predicting it. Leveraging the Transformer architecture, which incorporates positional information through positional encoding, each feature point can directly predict the endpoint coordinates $(x_s, y_s, x_e, y_e)$ of the line segment, rather than predicting the offsets of the endpoints from the centroid as in CNN-based methods \cite{huang2020tp, gu2022towards}.
\par
Given the processed feature point $\bm{p}$ in the feature map $\bm{F}$, we can get the confidence score $c$ indicating the confidence that the feature point should predict a line segment
\begin{equation}
    c = \sigma(\mathcal{F}_{c}(\bm{p})),
\end{equation}
and the location of the corresponding line segment
\begin{equation}
    \bm{l} = \mathcal{F}_{l}(\bm{p}),
\end{equation}
where $\mathcal{F}_{c}, \mathcal{F}_{l}$ are CNN-based prediction modules and $\sigma$ is the sigmoid function.

\subsection{Line Re-ranking Module}
While confidence scores are assigned to predicted line segments, these scores may not fully reflect the true quality of the segments since they are generated simultaneously. We propose optimizing these scores by evaluating the predicted line segments against corresponding geometric information, such as edges, endpoints, and length. To accomplish this, we perform sampling-based evaluations separately for endpoints and edges using the generated line segments. Specifically, we create endpoint maps and edge maps with a geometric information extractor. For a line segment to be considered high-quality, its endpoints should align with high-confidence regions on the endpoint map, and the segment should significantly overlap with high-confidence areas on the edge map.
\par
Given the endpoint map $\bm{M_e}$, we sample the endpoints $\bm{e_1}, \bm{e_2}$ according to the predicted line segments. The sampled scores are averaged as the endpoint confidence, which can be represented as
\begin{equation}
    s_e = \frac{1}{2}\sum\mathcal{S}(\{\bm{e_1}, \bm{e_2}\}; \bm{M_e})),
\end{equation}
$\mathcal{S}(\{\bm{e}\}; \bm{M})$ represents the sampling kernel indicating sampling $\bm{M}$ with the location $\bm{e}$.
Given the edge map $\bm{M_d}$, we first uniformly sample $m$ points between the two endpoints of the line segments. The scores are then sampled with these points from the edge map, and finally averaged as the edge score, which can be represented as
\begin{equation}
    s_d = \frac{1}{m}\sum\mathcal{S}(\{\frac{k}{m}\bm{e_1}+\frac{m-k}{m}\bm{e_2}\}_{k\in[1, m]}; \bm{M_d})).
\end{equation}
The geometric information extractor is composed of a group of CNNs in this work, which is supervised by ground truth edge maps and endpoint maps generated from the ground truth line segments.
\par
Moreover, to encourage the detection of long line segments and suppress the detection of fragmented line segments, the length is taken into account in evaluating line segment quality. To avoid the score expansion caused by length, the length scores is defined as
\begin{equation}
    s_l = ln(\|\bm{e_1}-\bm{e_2}\|_2+1).
\end{equation}
The optimized score is finally defined as the sum of the above terms and the previous confidence score $c$, which can be represented as
\begin{equation}
    s = \delta_{e}s_e + \delta_{d}s_d + \delta_{l}s_l + c.
\end{equation}
\par
By incorporating the re-ranking method, we can refine the confidence scores of predicted line segments by considering not only the centroid confidence score but also the overall quality of the line segments, as indicated by their endpoints, overlap with the edge map and the length.

\subsection{Network Architecture}
Our network architecture is built on a CNN-based backbone combined with a Deformable Transformer, designed specifically for line segment detection, which requires abundant low-level, high-resolution information. To capture features at various resolutions, we employ a feature pyramid during CNN feature extraction, feeding these multi-scale features into the Transformer. Notably, our approach simplifies the Transformer by using only the Encoder befitting from the stable line segment representation. Each feature token in the Encoder layers corresponds directly to a prediction target, eliminating the need for learnable queries and bipartite matching. For line segment prediction, we utilize the final $128\times128$ high-resolution features, which allow for finer predictions and reduce conflicts from overlapping centroids. Additionally, we enhance the network’s sensitivity to line segments by incorporating rotation augmentation during image feature extraction, rotating the images by ±90° and applying an inverse rotation afterward. Our overall architecture includes an image feature extractor based on ResNet50 with a feature pyramid and a feature processing module consisting of 6 deformable Transformer Encoder layers.

\begin{table*}[t]
\begin{center}
\setlength{\tabcolsep}{1mm}{
\begin{tabular}{|l|c c c c c|c c c c c|c|c|}
\hline
\multirow{2}{*}{Method} & \multicolumn{5}{c|}{Wireframe} & \multicolumn{5}{c|}{YUD} & \multirow{2}{*}{\makecell{Training \\ Epochs}} & \multirow{2}{*}{FPS} \\
& sAP$^{5}$ & sAP$^{10}$ & sF$^{10}$ & sF$^{15}$ & LAP & sAP$^{5}$ & sAP$^{10}$ & sF$^{10}$ & sF$^{15}$ & LAP & &  \\
\hline\hline
LSD \cite{von2008lsd} &6.7 &8.8 & - & - &18.7 & 7.5 & 9.2 & - & - & 16.1&- & 100.0\\
DWP \cite{huang2018learning} &3.7& 5.1 & - & - &6.6  &2.8 &2.6 & - & - &3.1&120 &2.2 \\
AFM \cite{xue2019learning} &18.3 &23.9 & - & - &36.7& 7.0& 9.1 & - & - &17.5 &200&14.1\\
LGNN \cite{meng2020lgnn} & - &62.3& - & - & - & - & - & - & - & - &-&15.8 \\
TP-LSD \cite{huang2020tp} &57.6 &57.2& - & - &61.3 &\textbf{27.6}&27.7& - & - &\textbf{34.3}& 400&20.0 \\ 
LETR \cite{xu2021line} &59.2 &65.6 & \underline{66.1} & \underline{67.4} &\underline{65.1} & 24.0 &27.6 & \underline{39.6} & \textbf{41.1} & 32.5 & 825 & 5.4\\
L-CNN \cite{zhou2019end}  &58.9 &62.8 & 61.3 & 62.4 &59.8& 25.9 &28.2 & 36.9 & 37.8 &32.0&16 &16.6\\
HAWP \cite{xue2020holistically} &62.5 &66.5 & 64.9 & 65.9 &62.9 &26.1 &\underline{28.5} & \textbf{39.7} & 40.5 &30.4 &30&32.9 \\
M-LSD \cite{gu2022towards} &56.4 &62.1 & - & - &61.5 &24.6 &27.3 & - & - &30.7 &150&115.4 \\ 
M-LSD\dag \cite{gu2022towards} &\underline{63.3} &\underline{67.1} & - & - &64.2 &\underline{27.5} &\underline{28.5} & - & - &32.4 & 150 & 32.9 \\ 
\hline\hline
RANK-LETR (Ours) & \textbf{65.0} & \textbf{69.7} & \textbf{66.7} & \textbf{67.7} & \textbf{65.6} & \textbf{27.6} & \textbf{30.1} & \textbf{39.7} & \underline{40.6} & \underline{34.1} & 120 & 9 \\ 
\hline
\end{tabular}}
\end{center}
\caption{Comparison results on Wireframe \cite{huang2018learning} and YUD \cite{denis2008efficient} datasets. We compare the proposed method with LSD \cite{von2008lsd}, DWP \cite{huang2018learning}, AFM \cite{xue2019learning}, LGNN \cite{meng2020lgnn}, TP-LSD\cite{huang2020tp}, LETR \cite{xu2021line}, L-CNN \cite{zhou2019end}, HAWP \cite{xue2020holistically} and M-LSD \cite{gu2022towards}. M-LSD\dag denotes the method of combining M-LSD and HAWP. Our method outperforms existing methods in prediction accuracy while requiring fewer training epochs than previous Transformer-based models.}
\label{tab:cmp}
\end{table*}

\begin{figure*}[t]
    \centering
    \includegraphics[width=17.7cm]{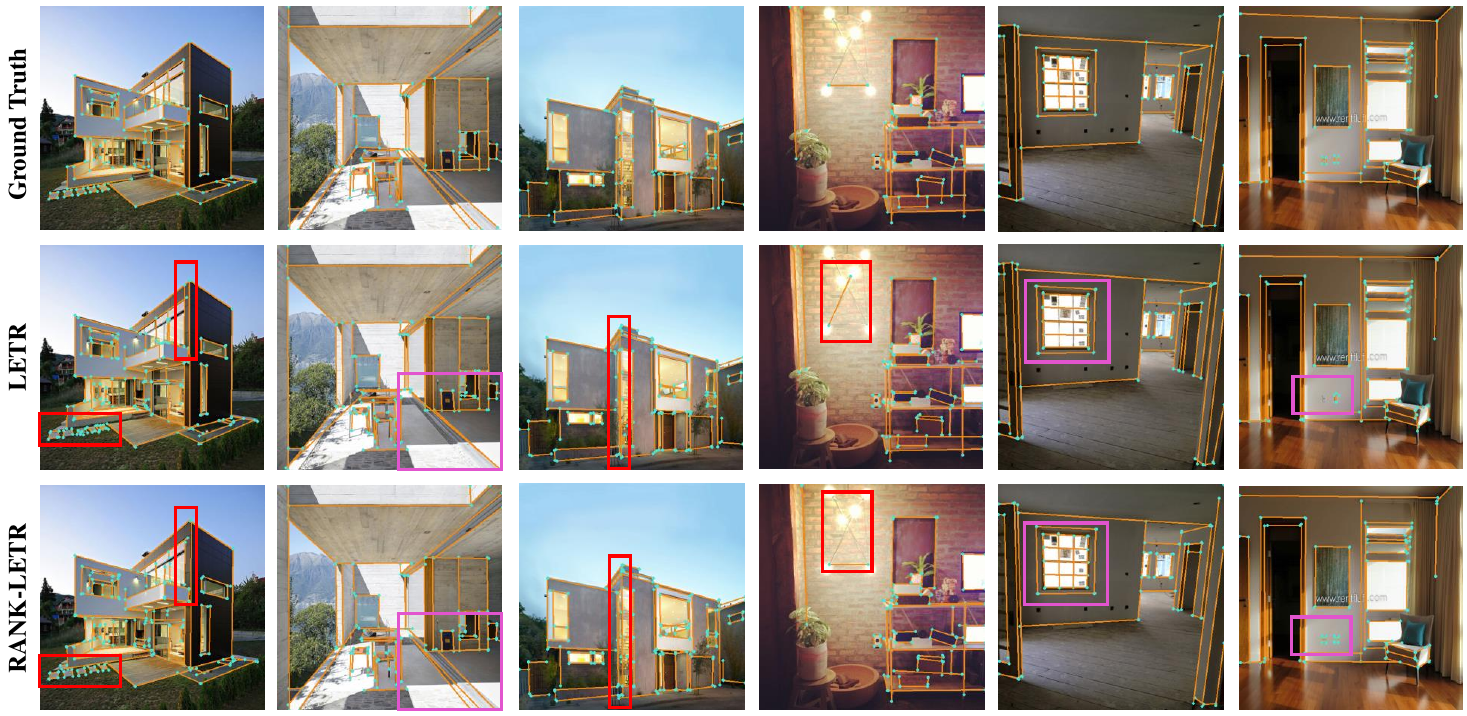}
    \caption{Visual examples of line segment detection results of two Transformer-based methods including LETR and ours on the Wireframe dataset. Our method can produce more accurate and complete detection results. For a better visual experience, we emphasize some examples of accurate detection with red bounding boxes and complete detection with purple bounding boxes.}
    \label{fig:showcmp}
\end{figure*}

\subsection{Training Supervision}
Since our method maps predicted results to ground truth before prediction, bipartite matching is no longer required to establish correspondences. Instead, we directly employ confidence loss and position loss to supervise our line segment detection model during training. Specifically, for confidence loss, we use binary cross-entropy loss to determine whether a feature point should predict a line segment, i.e., whether the centroid of the line segment is closest to the feature point. We find that the number of feature points that do not need to be responsible for predicting the line segments (defined as negative feature points) is considerably larger than those that need to be responsible (defined as positive feature points). Thus, we randomly select some of the former for training. The confidence loss $\mathcal{L}_{conf}$ is defined as

\begin{equation}
    \mathcal{L}_{conf}=-\frac{1}{N_{\pm}}\sum_{i=1}^{N_{\pm}}{(\hat{c_i}\log c_i+(1-\hat{c_i})\log(1-c_i))},
\end{equation}
where $N_{\pm}$ is the total number of the positive and selected negative feature points. $c_i$ is the predicted confidence score of the $i$-th feature point. $\hat{c_i}$ is the ground truth indicates that whether the $i$-th feature point should respond to detect a line segment and is annotated as $+1$ or $0$. 
\par
In the position loss, we minimize the L1 distance between the predicted line segment and ground truth ones on the image space. The loss is only applied to the positive feature points since the prediction of negative anchors should be meaningless. The presented position loss $\mathcal{L}_{pos}$ can be represented as
\begin{equation}
    \mathcal{L}_{pos}=\sum_{i=1}^{N_+}{\|\bm{l_i}-\bm{\hat{l_i}}\|_1},
\end{equation} 
where $N_+$ is the number of positive feature points. $\bm{l_i}$ represents a predicted line segments from the positive feature point and $\bm{\hat{l_i}}$ is the corresponding ground-truth.
\par
Inspired by the success of recent ranking-based losses in detection tasks, we introduce a line segment ranking loss to ensure that feature points with higher line segment detection quality receive higher confidence scores. Many ranking-based losses are guided by the quality of the bounding box or the predicting confidence. In the context of line segment detection, the natural basis for ranking is the quality of the predicted line segment. We define this quality using the Euclidean distance between the predicted line segment and its corresponding ground truth. A shorter distance indicates higher prediction quality. For all positive feature points, we perform pairwise comparisons based on their prediction quality, ensuring that feature points with higher prediction quality receive higher confidence scores, which can be represented as
\begin{equation}
\begin{aligned}
    \mathcal{L}_{rank}=&-\frac{1}{N_+^2}\sum_{i=1}^{N_+}\sum_{j=1}^{N_+}\sigma((c_j-c_i)) \\
    &(\|\bm{l_i}-\bm{\hat{l_i}}\|_2-\|\bm{l_j}-\bm{\hat{l_j}}\|_2),
\end{aligned}
\end{equation}
where we use sigmoid function to approximate the step function.
\par
Moreover, as we need to use the junction map and edge map in our line re-ranking module, we predict junction maps and edge maps from feature maps in different resolutions. For each resolution, we supervise the junction map $\mathcal{F}_j(\bm{F})$ and edge map $\mathcal{F}_e(\bm{F})$ as similar to ground truth maps $J$ and $E$, which can be represented as
\begin{equation}
    \mathcal{L}_{junc}=\sum_{i=1}^{n}{\|\mathcal{F}_j(\bm{F_i})-J_i\|_2},
\end{equation}
and
\begin{equation}
    \mathcal{L}_{edge}=\sum_{i=1}^{n}{\|\mathcal{F}_e(\bm{F_i})-E_i\|_2}.
\end{equation}
\par
The total loss can be defined as the sum of the above loss terms with proper weights, which can be written as
\begin{equation}
    \begin{aligned}
        \mathcal{L}_{total} = & \lambda_r \mathcal{L}_{rank} + \lambda_c \mathcal{L}_{conf} + \lambda_p \mathcal{L}_{pos} + \lambda_j \mathcal{L}_{junc} \\
        &+ \lambda_e \mathcal{L}_{edge}.
    \end{aligned}
\end{equation}

\section{Experimental Results}

In this section, we first describe the implementation setup of our method. Then we compare it with the state-of-the-art line segment detection approaches with quantitative experiments. In the ablation study, we demonstrate the effectiveness of the components in our method. We also conduct a parameter study to select better hyperparameters.

\subsection{Experimental Setup} 

\subsubsection{Datasets ans Metrics} 
We conduct our experiments in two publicly available datasets including the Wireframe dataset \cite{huang2018learning} and the YorkUrban dataset \cite{denis2008efficient}. The Wireframe dataset consists of 5,000 training and 462 test images of man-made environments, while the YorkUrban dataset has 102 test images. Following the typical training and test protocol \cite{huang2020tp, zhou2019end}, we train our model with the training set from the Wireframe dataset and test with both Wireframe and YorkUrban datasets. We evaluate our models using prevalent metrics for line segment detection task including structural average precision (sAP), the F-score measurement (sF) and line matching average precision (LAP).

\begin{table*}[tb]
    \centering
    \begin{tabular}{|c|c|c|c|c|c|c c c c|}
        \hline
         \multirow{2}{*}{\makecell{Matched \\ Predicting}} & \multirow{2}{*}{\makecell{\makecell{Rotation \\ augmentation}}} & \multirow{2}{*}{Re-ranking} & \multirow{2}{*}{\makecell{Resolution}} & \multirow{2}{*}{\makecell{Hidden \\ Dim}} & \multirow{2}{*}{\makecell{Referring \\ Points}} & \multicolumn{4}{c|}{Wireframe} \\
         & & & & & & sAP$^{5}$ & sAP$^{10}$ & sF$^{5}$ & sF$^{10}$ \\
        \hline\hline
         - & - & - & - & - & - & 50.3 & 58.5 & 56.8 & 61.5 \\
         \checkmark & - & - & 128 & 128 & 4 & 60.2 & 65.9 & 64.9 & 66.2\\
         \checkmark & \checkmark & - & 128 & 128 & 4 & 61.7 & 67.4 & 63.0 & 66.2 \\
         \checkmark & \checkmark & \checkmark & 64 & 128 & 4 & 63.4 & 68.3 & 62.7 & 65.5 \\
         \checkmark & \checkmark & \checkmark & 128 & 128 & 4 & \underline{65.0} & \underline{69.7} & \textbf{64.2} & \textbf{66.9} \\
        \hline\hline
         \checkmark & \checkmark & \checkmark & 128 & 64 & 4 & 62.9 & 68.3 & 62.7 & 65.8 \\
         \checkmark & \checkmark & \checkmark & 128 & 256 & 4 & \textbf{65.1} & \textbf{69.8} & \textbf{64.2} & \underline{66.8} \\
        \hline\hline
         \checkmark & \checkmark & \checkmark & 128 & 128 & 2 & 64.4 & 69.4 & 63.7 & 66.6\\
         \checkmark & \checkmark & \checkmark & 128 & 128 & 8 & 64.6 & 69.6 & 63.8 & 66.6 \\   
        \hline
    \end{tabular}
    \caption{Ablation and parameter study of our method on the Wireframe \cite{huang2018learning} dataset. We first construct a baseline method according to our modeling, and then gradually add different components. Experimental results show that applying matched-predicting strategy, re-ranking module, rotation augmentation and predicting on high-resolution feature maps can all boost the performance of our approach. We also select the number of hidden dim and referring points according to the parameter study on them.}
    \label{tab:ablation_architecture}
\end{table*}

\subsubsection{Implementation details} Our training and evaluation are implemented in PyTorch. We use 4 NVIDIA Tesla V100 cards for training and 1 card for evaluation. In training, We use AdamW as the model optimizer and set weight decay as $10^{-4}$. We train the model for $120$ epochs. The initial learning rates are set to $5 \times 10^{-4}$ for all parameters. Learning rates are reduced by a factor of 10 in epoch $60$ and $90$. We use a batch size of 8 and the size of the input images is set to $512\times512$. Similarly to many computer vision tasks, we adopt a CNN backbone pretrained on ImageNet, while other parameters are trained from scratch.
\par
The results of our method is predicted on the features of the resolution of $128\times128$. $\lambda_r, \lambda_c,\lambda_p,\lambda_j,\lambda_e$ are set to $1, 1, 10, 1, 1$, respectively. $m$ is set as 32 for sampling and $\delta_{e}, \delta_{d}, \delta_{l}$ are set to $0.5, 0.5, 0.5$ empirically for YUD dataset and $0.4, 0.1, 0.2$ after validing on random sampling set on training set. $500$ line segments with high scores are detected with NMS for the proposed method in all the experiences. 

\subsection{Comparison with the SOTA}

\begin{figure}[t]
    \centering
    \includegraphics[width=8.4cm]{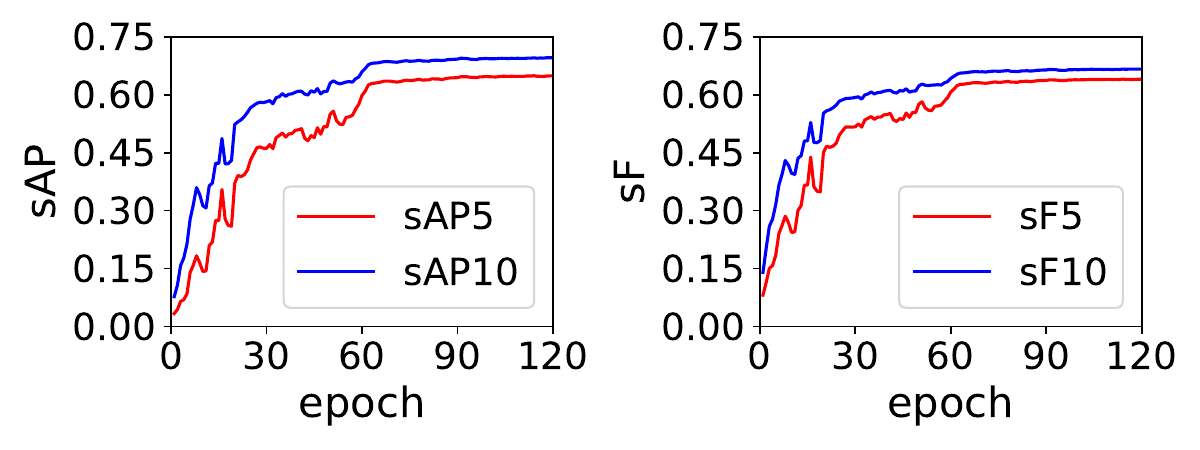}
    \caption{The accuracy curves for line segment detection over the time of training. We observed that our method reaches a high level of accuracy after just 60 epochs.}
    \label{fig:acc_epoch}
\end{figure}

Our comparisons are conducted on the Wireframe dataset \cite{huang2018learning} and the YorkUrban dataset \cite{denis2008efficient}.
We compare our method with the state-of-the-art methods including LSD \cite{von2008lsd}, DWP \cite{huang2018learning}, AFM \cite{xue2019learning}, LGNN \cite{meng2020lgnn}, TP-LSD \cite{huang2020tp}, LETR \cite{xu2021line}, L-CNN \cite{zhou2019end} , HAWP \cite{xue2020holistically} and M-LSD \cite{gu2022towards}. All the methods are learning-based methods except the classical LSD. The comparison results are listed in Table \ref{tab:cmp}. RANK-LETR outperforms existing Transformer-based and CNN-based methods in prediction accuracy, particularly on the Wireframe dataset. Additionally, our method can converge with fewer training epochs compared to previous Transformer-based models. We also show the accuracy curves for line segment detection over the time of training in Fig.\ref{fig:acc_epoch}. Our method reaches a high level of accuracy after just 60 epochs, which also proves that our method has a fast and stable convergence. 
\par
Visual examples of line segment detection results of two Transformer-based methods including LETR and ours on outdoor and indoor scenes in the Wireframe dataset are shown in Fig. \ref{fig:showcmp}. It is shown that our method can produce more accurate and complete detection result, especially in the detection of short line segments and in areas with high line density.
\par

\subsection{Ablation Study}

\begin{table}[t]
\begin{center}
\begin{tabular}{c|cccc}
\toprule
Method  & sAP$^{5}$ & sAP$^{10}$ & sF$^{5}$ & sF$^{10}$ \\
\midrule
w/o Ranking Loss & 26.5 & 29.2 & 36.9 & 38.9 \\
with Ranking Loss & \textbf{27.6} & \textbf{30.1} & \textbf{37.9} & \textbf{39.7} \\
\bottomrule
\end{tabular}
\end{center}
\caption{Line segment detection accuracy on YUD with and without ranking loss. We observe that the ranking loss can improve the generalization capability of the proposed method when trained and evaluated on different datasets.}
\label{tab:ranking_loss}
\end{table}

To verify the effectiveness of components and find the influence of the hyperparameters in our proposed method, we conduct an ablation and parameter study of our network architecture. The results are summarized in Table \ref{tab:ablation_architecture}. The ablation study is conducted on the Wireframe dataset and the \textit{sAP} and \textit{sF} results are reported. 
\par
We employed the combination of ResNet50 and Transformer Encoder and Decoder from the classical LETR method as our baseline approach. We constructed a foundational version of our modeling using a ResNet50 network and a 6-layer Deformable Transformer Encoder, where feature maps at resolutions of $128\times128$, $64\times64$, and $32\times32$ are input from the feature extraction backbone into the Transformer network for processing. Based on this architecture, we can use matched predicting strategy in supervising and find it can significantly improve the detection performance. On this basis, we sequentially added a re-ranking module and a rotation augmentation module, both of which further enhanced performance. Additionally, we compared the results of line segment prediction across different resolution feature maps, concluding that line segment prediction on high-resolution feature maps can provides a higher accuracy.
\par
\begin{figure}[t]
    \centering
    \includegraphics[width=8cm]{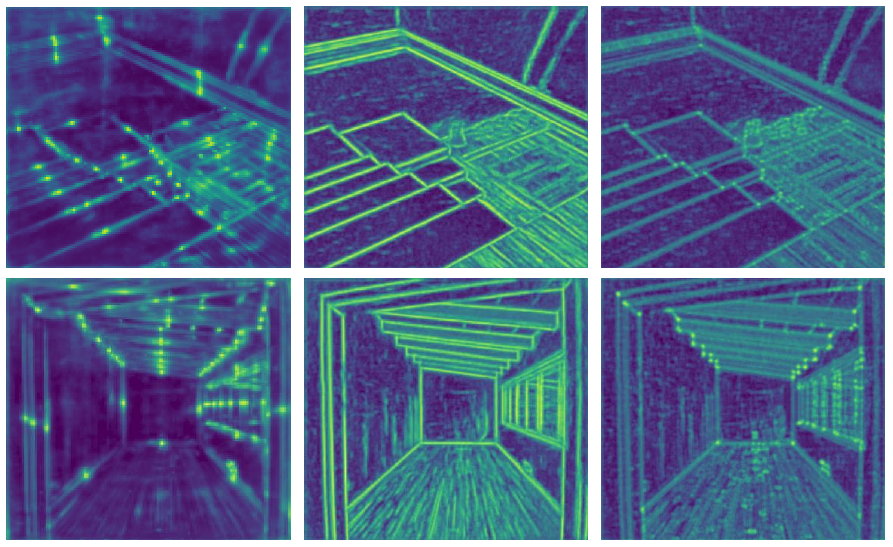}
    \caption{Saliency maps generated from score maps (left), edge maps (middle) and endpoint maps (right).}
    \label{fig:fm}
\end{figure}

Moreover, we vary the number of referring points and the hidden dim of the Transformer architecture to show the influence of these hyperparameters. We find the hidden dim may be appropriately set to more that $128$ since a smaller one will lead to a lower accuracy. Changing the number of referring points has little impact on detection performance. In the ablation studies above, except for the baseline method that required 240 epochs to ensure convergence, all other methods were trained for 120 epochs, demonstrating the improved convergence capabilities of our proposed strategy. The saliency maps generated from some feature maps are also presented in Fig.\ref{fig:fm}
to exhibit the learning effect. We also demonstrate the effectiveness of the ranking loss in the Table \ref{tab:ranking_loss}. We found that using the ranking loss can improve the performance of the model when trained on the Wireframe dataset and tested on the YUD dataset. This indicates that the ranking loss can enhance the generalization performance of the model.
\par 

While our method generally achieves the highest detection accuracy, it has a limitation when the centroids of two line segments are very close. In such cases, our method can only predict one of the segments. Although we find that nearby grid points may predict the corresponding line segment instead of the grid where the centroid lies, this remains a theoretical limitation of our approach. Allowing each feature point to predict more than one, e.g., two line segments could potentially mitigate this issue, which we leave as future work.

\section{Conclusion}
We identify a critical factor limiting the performance of proposal-based line segment detection algorithms: the confidence scores may fail to accurately reflect the quality of the predicted line segments. To tackle this issue, we propose three novel techniques to enhance Transformer-based line segment detection methods. First, we introduce a simple and efficient line re-ranking module that optimizes the confidence scores of lines using learnable geometric information, such as edges, endpoints, and line lengths. This module is more interpretable and allows for flexible weighting across different scenarios. We also present a matched prediction strategy, wherein each feature point is responsible for detecting the line segment whose centroid is closest to it. Furthermore, we propose a line segment ranking loss to make feature points predict higher confidence scores for higher-quality predicted line segments during training. Building on these techniques, we developed a novel line segment detection model named RANK-LETR, which outperforms existing Transformer-based and CNN-based methods while requiring fewer training epochs than previous Transformer-based models.

\bibliography{aaai25}

\end{document}